\documentclass[conference]{IEEEtran} 
\ifCLASSOPTIONcompsoc
  \usepackage[nocompress]{cite}
\else
  \usepackage{cite}
\fi
\ifCLASSINFOpdf
\else
\fi
\usepackage[colorlinks=true,
            linkcolor=red,
            urlcolor=blue,
            citecolor=blue]{hyperref}
\usepackage{amsmath}
\usepackage{graphicx}
\usepackage{mathtools}
\usepackage{relsize}
\usepackage{graphicx}
\usepackage{algorithm}
\usepackage{filecontents,lipsum}
\usepackage{cite}
\usepackage{comment}
\usepackage{multirow}
\usepackage{amsfonts}
\usepackage[noend]{algpseudocode}
\usepackage[utf8]{inputenc}
\usepackage{graphicx}
\usepackage{caption}
\usepackage{latexsym}
\usepackage{etoolbox}
\usepackage{subcaption, float}
\makeatletter
\usepackage[inline]{enumitem}
\def\BState{\State\hskip-\ALG@thistlm}
\makeatother
\newcommand\given[1][]{\:#1\vert\:}
\AfterEndEnvironment{table}{\vskip-1ex}
\begin{document}

%
\title{HDLTex: \underline{H}ierarchical \underline{D}eep \underline{L}earning for \underline{Tex}t Classification}


\author{\IEEEauthorblockN{Kamran Kowsari\IEEEauthorrefmark{1},
Donald E. Brown\IEEEauthorrefmark{4}\IEEEauthorrefmark{3},
 Mojtaba Heidarysafa\IEEEauthorrefmark{4}, \\
 Kiana Jafari Meimandi\IEEEauthorrefmark{4},
Matthew S. Gerber\IEEEauthorrefmark{4}\IEEEauthorrefmark{3}, and
Laura E. Barnes\IEEEauthorrefmark{4}\IEEEauthorrefmark{3}}

\IEEEauthorblockA{\IEEEauthorrefmark{1}Department of Computer Science,
University of Virginia,
Charlottesville, VA, USA\\}
\IEEEauthorblockA{\IEEEauthorrefmark{4} Department of System and Information Engineering,
University of Virginia,
Charlottesville, VA, USA}

\IEEEauthorblockA{\IEEEauthorrefmark{3} Data Science Institute, 
University of Virginia,
Charlottesville, VA, USA}

\{\href{mailto:kk7nc@virginia.edu}{kk7nc},
\href{mailto:deb@virginia.edu}{deb},
\href{mailto:mh4pk@virginia.edu}{mh4pk}, 
\href{mailto:kj6vd@virginia.edu}{kj6vd}, 
\href{mailto:msg8u@virginia.edu}{msg8u}, 
\href{mailto:lbarnes@virginia.edu}{lbarnes}\}@virginia.edu}


%


\maketitle
\begin{abstract}
Increasingly large document collections require improved information processing methods for searching, retrieving, and organizing  text. Central to these information processing methods is document classification, which has become an important application for supervised learning. Recently the performance of traditional supervised classifiers has degraded as the number of documents has increased. This is because along with growth in the number of documents has come an increase in the number of categories. This paper approaches this problem differently from current document classification methods that view the problem as multi-class classification. Instead we perform hierarchical classification using an approach we call Hierarchical Deep Learning for Text classification~(HDLTex). HDLTex employs stacks of deep learning architectures to provide specialized understanding at each level of the document hierarchy.\\

\end{abstract}

\begin{IEEEkeywords} 
Text Mining; Document Classification; Deep Neural Networks; Hierarchical Learning; Deep Learning 
\end{IEEEkeywords}

\IEEEpeerreviewmaketitle
\section{Introduction}\label{sec:intro}
Each year scientific researchers produce a massive number of documents. In 2014 the 28,100 active, scholarly, peer-reviewed, English-language journals published about 2.5 million articles, and there is evidence that the rate of growth in both new journals and publications is accelerating~\cite{ware2015stm}. The volume of these documents has made automatic organization and classification an essential element for the advancement of basic and applied research. Much of the recent work on automatic document classification has involved supervised learning techniques such as classification trees, na\"{i}ve Bayes, support vector machines (SVM), neural nets, and ensemble methods. Classification trees and na\"{i}ve Bayes approaches provide good interpretability but tend to be less accurate than the other methods. 

However, automatic classification has become increasingly challenging over the last several years due to growth in corpus sizes and the number of fields and sub-fields. Areas of research that were little known only five years ago have now become areas of high growth and interest. This growth in sub-fields has occurred across a range of disciplines including biology (e.g., CRISPR-CA9), material science (e.g., chemical programming), and health sciences (e.g., precision medicine). This growth in sub-fields means that it is important to not just label a document by specialized area but to also organize it within its overall field and the accompanying sub-field. This is hierarchical classification.

Although many existing approaches to document classification can quickly identify the overall area of a document, few of them can rapidly organize documents into the correct sub-fields or areas of specialization. Further, the combination of top-level fields and all sub-fields presents current document classification approaches with a combinatorially increasing number of class labels that they cannot handle. This paper presents a new approach to hierarchical document classification that we call Hierarchical Deep Learning for Text classification~(HDLTex).\footnote{HDLTex is shared as an open source tool at \url{https://github.com/kk7nc/HDLTex}} HDLTex combines deep learning architectures to allow both overall and specialized learning by level of the document hierarchy. This paper reports our experiments with HDLTex, which exhibits improved accuracy over traditional document classification methods. 

\section{Related Work}\label{sec:related}

Document classification is necessary to organize documents for retrieval, analysis, curation, and annotation. Researchers have studied and developed a variety of methods for document classification. Work in the information retrieval community has focused on search engine fundamentals such as indexing and dictionaries that are considered core technologies in this field~\cite{manning2008introduction}. Considerable work has built on these foundational methods to provide improvements through feedback and query reformulation~\cite{french1997classification,kowsari2015construction}. 

More recent work has employed methods from data mining and machine learning. Among the most accurate of these techniques is the support vector machine (SVM)~\cite{joachims1999transductive,tong2001support,fernandez2014we}. SVMs use kernel functions to find separating hyperplanes in high-dimensional spaces. Other kernel methods used for information retrieval include string kernels such as the spectrum kernel~\cite{leslie2002spectrum} and the mismatch kernel~\cite{eskin2003mismatch}, which are widely used with DNA and RNA sequence data.

SVM and related methods are difficult to interpret. For this reason many information retrieval systems use decision trees~\cite{french1997classification} and na\"{i}ve Bayes~\cite{mccallum1998comparison,kim2006some} methods. These methods are easier to understand and, as such, can support query reformulation, but they lack accuracy. Some recent work has investigated topic modeling to provide similar interpretations as na\"{i}ve Bayes methods but with improved accuracy~\cite{van2017effective}.

This paper uses newer methods of machine learning for document classification taken from deep learning. Deep learning is an efficient version of neural networks~\cite{hinton2006reducing} that can perform unsupervised, supervised, and semi-supervised learning~\cite{johnson2014effective}. Deep learning has been extensively used for image processing, but many recent studies have applied deep learning in other domains such as text and data mining. The basic architecture in a neural network is a fully connected network of nonlinear processing nodes organized as layers. The first layer is the input layer, the final layer is the output layer, and all other layers are hidden. In this paper, we will refer to these fully connected networks as Deep Neural Networks (DNN). Convolutional Neural Networks (CNNs) are modeled after the architecture of the visual cortex where neurons are not fully connected but are spatially distinct~\cite{lecun1998gradient}. CNNs provide excellent results in generalizing the classification of objects in images~\cite{oquab2014learning}.  More recent work has used CNNs for text mining~\cite{lee2016sequential}. In research closely related to the work in this paper, Zhang et al.~\cite{zhang2015character} used CNNs for text classification with character-level features provided by a fully connected DNN. Regardless of the application, CNNs require large training sets. Another fundamental deep learning architecture used in this paper is the Recurrent Neural Network (RNN). RNNs connect the output of a layer back to its input. This architecture is particularly important for learning time-dependent structures to include words or characters in text~\cite{medsker2001recurrent}. 
Deep learning for hierarchical classification is not new with this paper, although the specific architectures, the comparative analyses, and the application to document classification are new. Salakhutdinov~\cite{salakhutdinov2013learning,huang2012learning} used deep learning to hierarchically categorize images. At the top level the images are labeled as animals or vehicles. The next level then classifies the kind of animal or vehicle. This paper describes the use of deep learning approaches to create a hierarchical document classification approach. These deep learning methods have the promise of providing greater accuracy than SVM and related methods. Deep learning methods also provide flexible architectures that we have used to produce hierarchical classifications. The hierarchical classification our methods produce is not only highly accurate but also enables greater understanding of the resulting classification by showing where the document sits within a field or area of study.

\begin{figure*}
\centering
\includegraphics[width=\textwidth]{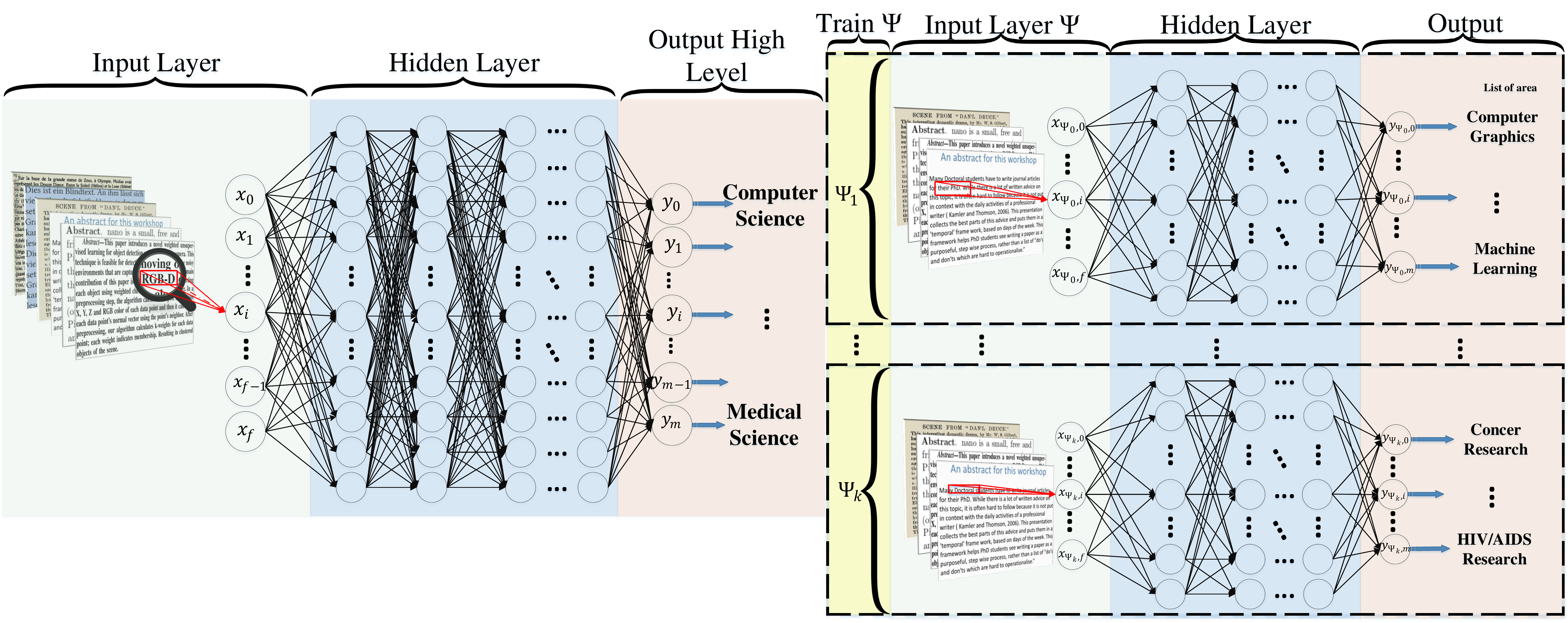}
\caption{HDLTex: \underline{H}ierarchical \underline{D}eep \underline{L}earning for \underline{Tex}t Classification. This is our Deep Neural Network~(DNN) approach for text classification. The left figure depicts the parent-level of our model, and the right figure depicts child-level models defined by $\Psi_i$ as input documents in the parent level.}\label{Fig_DNN}
\end{figure*}

\section{Baseline Techniques}\label{sec:me_ba}

This paper compares fifteen methods for performing document classification. Six of these methods are baselines since they are used for traditional, non-hierarchical document classification. Of the six baseline methods three are widely used for document classification: term-weighted support vector machines~\cite{chen2016turning}, multi-word support vector machines\cite{zhang2008text},  and na\"{i}ve Bayes classification~(NBC). The other three are newer deep learning methods that form the basis for our implementation of a new approach for hierarchical document classification.  These deep learning methods are described in Section~\ref{sec:deep_learn}.

\subsection{Support Vector Machines~(SVMs)}\label{subsec:me_svm}

Vapnik and Chervonenkis introduced the SVM in 1963 \cite{vapnik1964class,chervonenkis2013early}, and in 1992 Boser et al. introduced a nonlinear version to address more complex classification problems~\cite{boser1992training}. The key idea of the nonlinear SVM is the generating kernel shown in Equation \ref{SVM_kernel}, followed by Equations~\ref{SVM_kernel2} and~\ref{SVM_kernel3}:
\begin{align} 
\label{SVM_kernel}
&K(x,x') = < \phi (x) , \phi (x') > \\
\label{SVM_kernel2}
&f(x) =  \sum_{x_i \in training} \alpha_i K(x,x_i) + b \\
\begin{split} 
\label{SVM_kernel3}
&\max_{\alpha_1,...,\alpha_n}~\sum_{i=1}^{n} \alpha_i - \frac{1}{2} \sum_{j=1}^n \sum_{k=1}^n \alpha_j \alpha_k y_j y_k K(x_j,x_k)\\
\forall& \alpha_i \geq 0 i \in {1,.., n}.
\end{split}
\end{align}

\subsubsection*{Multi-Class SVM} Text classification using string kernels within SVMs has been successful in many research projects~\cite{singh2017gakco}. The original SVM solves a binary classification problem; however, since document classification often involves several classes, the binary SVM requires an extension. In general, the multi-class SVM (MSVM) solves the following optimization:
\begin{equation}
\min_{w_1,w_2,..,w_k,\zeta}~\frac{1}{2}\sum_{k}w_k^Tw_k + C\sum_{(x_i,y_i) \in D} \zeta_i
\end{equation}
\begin{align}
\begin{split}
st.~~ &w_{y_i}^Tx - w_k^Tx \leq i - \zeta_i,~~\\
&\forall (x_i,y_i) \in D, k\in\{1,2,...,K\}, k\neq y_i
\end{split}
\end{align} 
where $k$ indicates number of classes, $\zeta_i$ are slack variables, and $w$ is the learning parameter. To solve the MSVM we construct a decision function of all $k$ classes at once \cite{chen2016turning,weston1998multi}. One approach to MSVM is to use binary SVM to compare each of the $k(k-1)$ pairwise classification labels, where $k$ is the number of labels or classes. Yet another technique for MSVM is one-versus-all, where the two classes are one of the $k$ labels versus all of the other~$k-1$ labels.
\subsubsection*{Stacking Support Vector Machines~(SVM)}
We use Stacking SVMs as another baseline method for comparison with HDLTex. The stacking SVM provides an ensemble of individual SVM classifiers and generally produces more accurate results than single-SVM models  ~\cite{sun2001hierarchical,sebastiani2002machine}.

\begin{figure*}[t]
 \centering
\includegraphics[width=\textwidth]{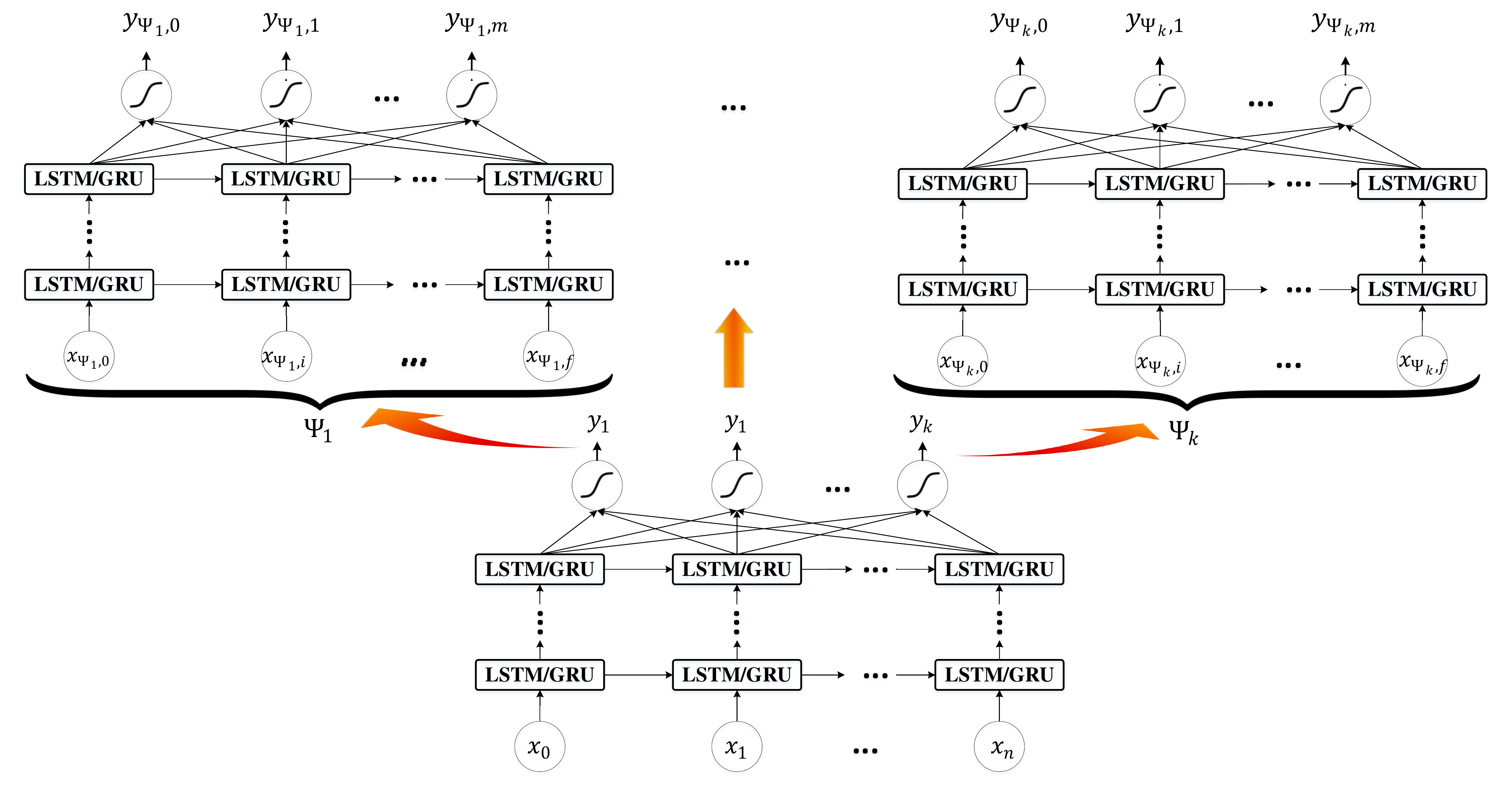}
\caption{HDLTex: \underline{H}ierarchical \underline{D}eep \underline{L}earning for \underline{Tex}t Classification. This is our structure of recurrent neural networks~(RNN) for text classification. The left figure is the parent level of our text leaning model. The right figure depicts child-level learning models defined by $\Psi_i$ as input documents in the parent levels.
}\label{Fig_RNN}
\end{figure*}

\subsection{Na\"{i}ve Bayes classification}\label{subsec:me_NB}

Na\"{i}ve Bayes is a simple supervised learning technique often used for information retrieval due to its speed and interpretability~\cite{lewis1998naive,manning2008introduction}. Suppose the number of  documents is $n$ and each document has the label $c, c \in \{ c_1,c_2,...,c_k\}$, where $k$ is the number of labels. Na\"{i}ve Bayes calculates
\begin{align}
P(c \given d) = \frac{P(d \given c)P(c)}{P(d)}
\end{align}
where $d$ is the document, resulting in
\begin{align} \label{eq1}
\begin{split}
C_{MAP} = &~arg \max_{c \in C} P(d \given c)P(c) \\
        = &~arg \max_{c\in C} P(x_1,x_2,...,x_n \given c)P(c).
\end{split}
\end{align}
The na\"{i}ve Bayes document classifier used for this study uses word-level classification~\cite{kim2006some}. Let $\hat{\theta}_j$ be the parameter for word $j$, then
\begin{equation}
    P(c_j \given d_i;\hat{\theta}) = \frac{P(c_j \given \hat{\theta})P(d_i \given c_j ; \hat{\theta}_j)}{P(d_i \given \hat{\theta})}.
\end{equation}


\section{Feature Extraction}\label{sec:feature_extract}

Documents enter our hierarchical models via features extracted from the text. We employed different feature extraction approaches for the deep learning architectures we built. For CNN and RNN, we used the text vector-space models using $100$ dimensions as described in Glove~\cite{pennington2014glove}. A vector-space model is a mathematical mapping of the word space, defined as
\begin{equation}
    d_j = (w_{1,j},w_{2,j},...,w_{i,j}...,w_{l_j,j})
\end{equation}
where $l_j$ is the length of the document $j$, and $w_{i,j}$ is the Glove word embedding vectorization of word $i$ in document $j$.

For DNN, we used count-based and term frequency–inverse document frequency~(tf-idf) for feature extraction. This approach uses counts for $N$-grams, which are sequences of $N$ words~\cite{cavnar1994n,glickman1985comparing}. For example, the text ``In this paper we introduced this technique'' is composed of the following N-grams:
\begin{itemize}
\item Feature count (1):  \{ (In, 1) , (this, 2), (paper, 1), (we, 1), (introduced, 1), (technique, 1) \}
\item Feature count (2):  \{ (In, 1) , (this, 2), (paper,~1), (we,~1), (introduced, 1), (technique, 1), (In this, 1), (This paper,~1), (paper we,~1),...\}
\end{itemize}
Where the counts are indexed by the maximum $N$-grams. So Feature count (2) includes both 1-grams and 2-grams. The resulting DNN feature space is
\begin{align}
\begin{split}
     f_{j,n}=& [x_{(j,0)},...,x_{(j,k-1)},x_{j,\{0,1\}},
     ...,\\&x_{j,\{k-2,k-1\}},...,x_{j,\{k-n,...,k-1\}}]   
\end{split}
\end{align}
where $f$ is the feature space of document $j$ for $n$-grams of size $n, n \in \{0,1,...,N\}$, and $x$ is determined by word or $n$-gram counts. Our algorithm is able to use N-grams for features within deep learning models \cite{kevselj2003n}.

\section{Deep Learning Neural Networks}\label{sec:deep_learn}

The methods used in this paper extend the concepts of deep learning neural networks to the hierarchical document classification problem. Deep learning neural networks provide efficient computational models using combinations of non-linear processing elements organized in layers. This organization of simple elements allows the total network to generalize~(i.e., predict correctly on new data)~\cite{lecun2015deep}. In the research described here, we used several different deep learning techniques and combinations of these techniques to create hierarchical document classifiers. The following subsections provide an overview of the three deep learning architectures we used: Deep Neural Networks~(DNN), Recurrent Neural Networks(RNN), and Convolutional Neural Networks~(CNN).

\subsection{Deep Neural Networks (DNN)}\label{subsec:DNN}
In the DNN architecture each layer only receives input from the previous layer and outputs to the next layer. The layers are fully connected. The input layer consists of the text features (see~\ref{sec:feature_extract}) and the output layer has a node for each classification label or only one node if it is a binary classification. This architecture is the baseline DNN. Additional details on this architecture can be found in~\cite{lanchantin2016deep}.

This paper extends this baseline architecture to allow hierarchical classification. Figure~\ref{Fig_DNN} shows this new architecture. The DNN for the first level of classification (on the left side in Figure~\ref{Fig_DNN}) is the same as the baseline DNN. The second level classification in the hierarchy consists of a DNN trained for the domain output in the first hierarchical level. Each second level in the DNN is connected to the output of the first level. For example, if the output of the first model is labeled \textit{computer science} then the DNN in the next hierarchical level~(e.g., $\Psi_1$ in Figure~\ref{Fig_DNN}) is trained only with all \textit{computer science} documents. So while the first hierarchical level DNN is trained with all documents, each DNN in the next level of the document hierarchy is trained only with the documents for the specified domain.

The DNNs in this study are trained with the standard back-propagation algorithm using both sigmoid (Equation~\ref{sigmoid}) and ReLU (Equation~\ref{relu}) as activation functions. The output layer uses softmax (Equation~\ref{Softmax}).
\begin{align}
\label{sigmoid}
f(x) =& \frac{1}{1+e^{-x}}\in (0,1),\\
\label{relu}
f(x) =& \max(0,x),\\
\label{Softmax}
\sigma(z)_j=& \frac{e^{z_j}}{\sum_{k=1}^K e^{z_k}},\\
\forall &  ~j \in \{1, …, K\} \nonumber
\end{align}
Given a set of example pairs $(x,y), x \in X, y \in Y$ the goal is to learn from the input and target spaces using hidden layers. In text classification, the input is generated by vectorization of text (see Section~\ref{sec:feature_extract}).


\subsection{Recurrent Neural Networks~(RNN)}\label{subsec:RNN}

The second deep learning neural network architecture we use is RNN. In RNN the output from a layer of nodes can reenter as input to that layer. This approach has advantages for text processing~\cite{KarpathyRNN}. The general RNN formulation is given in Equation~\ref{rnn_gen} where $x_t$ is the state at time $t$ and $\boldsymbol{u_t}$ refers to the input at step $t$.
\begin{equation}
\label{rnn_gen}
x_{t}=F(x_{t-1},\boldsymbol{u_t},\theta)
\end{equation}
We use weights to reformulate Equation~\ref{rnn_gen} as shown in Equation~\ref{rnn_spec} below:
\begin{equation}\label{rnn_spec}
x_{t}=\mathbf{W_{rec}}\sigma(x_{t-1})+\mathbf{W_{in}}\mathbf{u_t}+\mathbf{b}.
\end{equation}
In Equation~\ref{rnn_spec}, $\mathbf{W_{rec}}$ is the recurrent matrix weight, $\mathbf{W_{in}}$ are the input weights, $\mathbf{b}$ is the bias, and $\sigma$ is an element-wise function. Again we have modified the basic architecture for use in hierarchical classification. Figure~\ref{Fig_RNN} shows this extended RNN architecture.

Several problems (e.g., vanishing and exploding gradients) arise in RNNs when the error of the gradient descent algorithm is back-propagated through the network~\cite{bengio1994learning}.  To deal with these problems, long short-term memory~(LSTM) is a special type of RNN that preserves long-term dependencies in a more effective way compared with the basic RNN. This is particularly effective at mitigating the vanishing gradient problem~\cite{pascanu2013difficulty}. 
\begin{figure}[H]
\centering
\includegraphics[width=0.77\columnwidth]{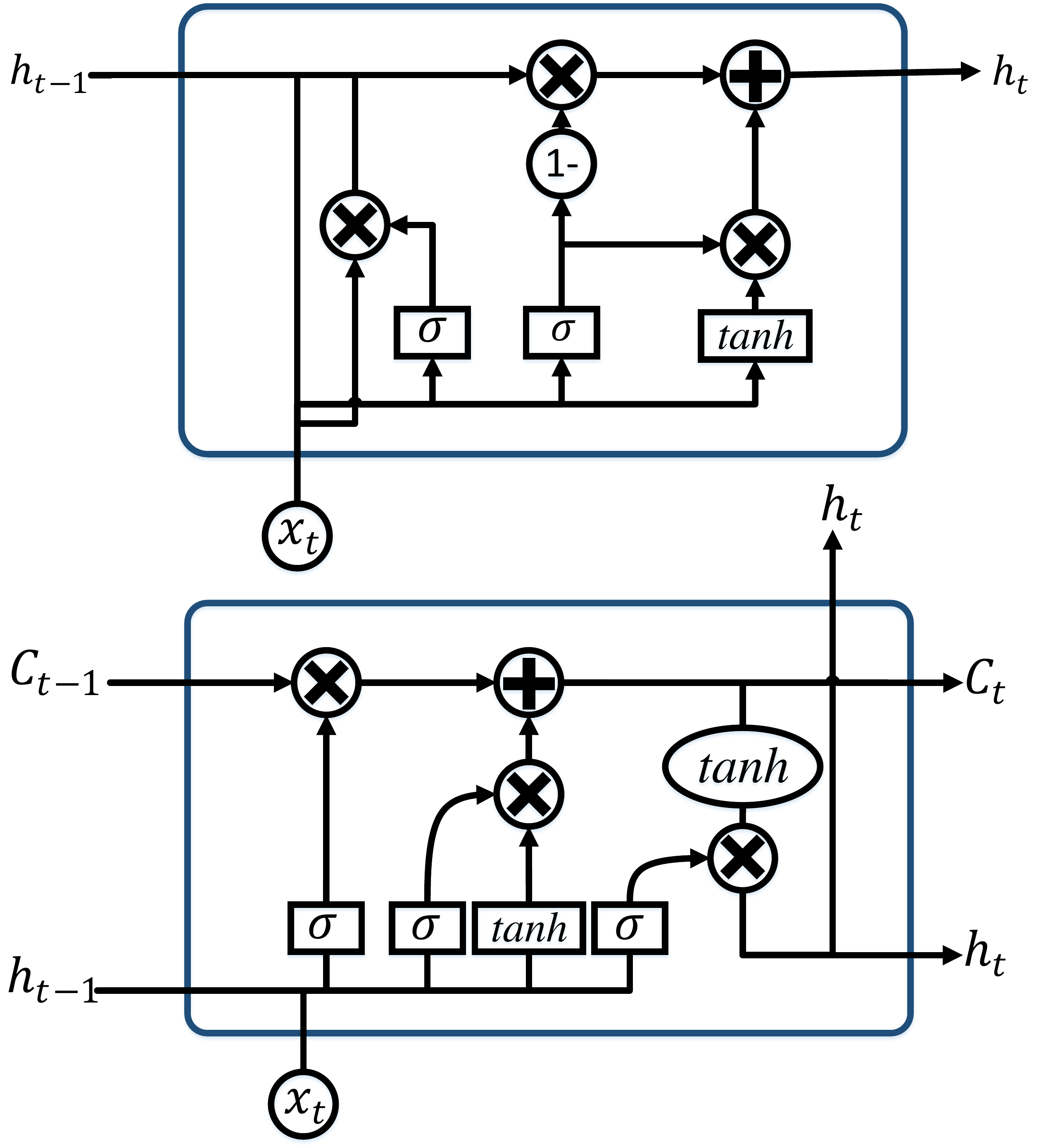}
\caption{The top sub-figure is a cell of GRU, and the bottom Figure is a cell of LSTM.}\label{fig:LSTM}
\end{figure}
Figure~\ref{fig:LSTM} shows the basic cell of an LSTM model. Although LSTM has a chain-like structure similar to RNN, LSTM uses multiple gates to regulate the amount of information allowed into each node state. A step-by-step explanation the LSTM cell and its gates is provided below:
\begin{enumerate}
    \item Input Gate:
    \begin{align}
    i_{t}=\sigma(W_{i}[x_{t},h_{t-1}]+b_{i}),
    \end{align}
    \item Candid Memory Cell Value:
    \begin{align}
    \tilde{C_{t}}=\tanh(W_{c}[x_{t},h_{t-1}]+b_{c}),
    \end{align}
    \item Forget Gate Activation:
    \begin{align}
    f_{t}=\sigma(W_{f}[x_{t},h_{t-1}]+b_{f}),
    \end{align}
    \item New Memory Cell Value:
    \begin{align}
    C_{t}=  i_{t}* \tilde{C_{t}}+f_{t} C_{t-1},
    \end{align}
    \item Output Gate Values:
    \begin{align}
    \left. \begin{aligned}
    o_{t}=& \sigma(W_{o}[x_{t},h_{t-1}]+b_{o}),\\
    h_{t}=&o_{t}\tanh(C_{t}),
    \end{aligned}\right.
    \end{align}
\end{enumerate}

In the above description, $b$ is a bias vector, $W$ is a weight matrix, and $x_{t}$ is the input to the memory cell at time~$t$. The~$i,c,f$ and $o$ indices refer to input, cell memory, forget and output gates, respectively. Figure~\ref{fig:LSTM} shows the structure of these gates with a graphical representation.

An RNN can be biased when later words are more influential than the earlier ones. To overcome this bias convolutional neural network~(CNN) models~(discussed in Section~\ref{subsec:CNN}) include a max-pooling layer to determine discriminative phrases in text~\cite{lai2015recurrent}. A gated recurrent unit~(GRU) is a gating mechanism for RNNs that was introduced in 2014~\cite{cho2014learning}. GRU is a simplified variant of the LSTM architecture, but there are differences as follows: GRUs contain two gates, they do not possess internal memory~(the~$C_{t-1}$ in Figure~\ref{fig:LSTM}), and a second non-linearity is not applied~($tanh$ in Figure~\ref{fig:LSTM}).

\subsection{Convolutional Neural Networks~(CNN)}\label{subsec:CNN}
The final deep learning approach we developed for hierarchical document classification is the convolutional neural network~(CNN). Although originally built for image processing, as discussed in Section~\ref{sec:related}, CNNs have also been effectively used for text classification ~\cite{lecun1998gradient}. The basic convolutional layer in a CNN connects to a small subset of the inputs usually of size~$3 \times 3$. Similarly the next convolutional layer connects to only a subset of its preceding layer. In this way these convolution layers, called feature maps, can be stacked to provide multiple filters on the input. To reduce computational complexity, CNNs use pooling to reduce the size of the output from one stack of layers to the next in the network. Different pooling techniques are used to reduce outputs while preserving important features~\cite{scherer2010evaluation}. The most common pooling method is max-pooling where the maximum element is selected in the pooling window. In order to feed the pooled output from stacked featured maps to the next layer, the maps are flattened into one column. The final layers in a CNN are typically fully connected. In general during the back-propagation step of a CNN not only the weights are adjusted but also the feature detector filters. A potential problem of CNNs used for text is the number of channels or size of the feature space. This might be very large~(e.g., 50K words) for text, but for images this is less of a problem~(e.g., only 3 channels of RGB)~\cite{johnson2014effective}.

\subsection{Hierarchical Deep Learning}\label{subsec:HDL}

The primary contribution of this research is hierarchical classification of documents.  A traditional multi-class classification technique can work well for a limited number classes, but performance drops with increasing number of classes, as is present in hierarchically organized documents. In our hierarchical deep learning model we solve this problem by creating architectures that specialize deep learning approaches for their level of the document hierarchy~(e.g., see Figure~\ref{Fig_DNN}).  The structure of our Hierarchical Deep Learning for Text (HDLTex) architecture for each deep learning model is as follows:
\begin{description}
\item [DNN:] $8$ hidden layers with $1024$ cells in each hidden layer.
\item [RNN:] GRU and LSTM are used in this implementation, $100$ cells with GRU with two hidden layers.
\item [CNN:] Filter sizes of $\{3, 4, 5, 6, 7\}$ and max-pool of $5$, layer sizes of $\{128,128,128\}$ with max pooling of $\{5,5,35\}$, the CNN contains $8$ hidden layers.
\end{description}
All models used the following parameters: \textit{Batch Size = 128}, \textit{learning parameters = $0.001$, $\beta_1$=0.9, $\beta_2$=0.999, $\epsilon=1e^{08}$,   $decay=0.0$,  Dropout=0.5~(DNN) and Dropout=0.25~(CNN and RNN)}.

\subsection{Evaluation}\label{subsec:eval}
We used the following cost function for the deep learning models:
\begin{align}
\label{eq_cost}
\begin{split}
Acc(X) =&\sum_{\varrho} \bigg{[} \frac{Acc(X_{\Psi_\varrho})}{k_{\varrho}-1}\\& \sum_{\Psi \in \{ \Psi_1,..\Psi_k\}} Acc(X_{\Psi_i}).n_{\Psi_k}\bigg{]}
\end{split}
\end{align}
where $\varrho$ is the number of levels, $k$ indicates number of classes for each level, and $\Psi$ refers to the number of classes in the child's level of the hierarchical model. 

\subsection{Optimization}\label{subsec:opt}
We used two types of stochastic gradient optimization for the deep learning models in this paper: RMSProp and Adam. These are described below.

\subsubsection*{RMSProp Optimizer}

The basic stochastic gradient descent~(SGD) is shown below:
\begin{align}
\label{SGD}
\theta \leftarrow \theta - \alpha \nabla_\theta J(\theta , x_i,y_i)\\
\label{momentum}
\theta \leftarrow \theta -\big( \gamma \theta + \alpha \nabla_\theta J(\theta , x_i,y_i)\big)
\end{align}
For these equations, $\theta$ is the learning parameter, $\alpha$ is the learning rate, and $J(\theta , x_i,y_i)$ is the objective or cost function. The history of updates is defined by $\gamma \in (0,1)$. To update parameters, SGD uses a momentum term on a rescaled gradient, which is shown in Equation~(\ref{momentum}). This approach to the optimization does not perform bias correction, which is a problem for a sparse gradient.
 
\subsubsection*{Adam Optimizer}
Adam is another stochastic gradient optimizer, which averages over only the first two moments of the gradient, $v$ and $m$, as shown below:
\begin{align}
\label{adam}
\theta  \leftarrow & \theta - \frac{\alpha}{\sqrt{\hat{v}}+\epsilon} \hat{m}\\
\textit{where}\nonumber\\
\label{adam1}
g_{i,t} &=  \nabla_\theta J(\theta_i , x_i,y_i) \\
\label{adam2}
m_t &= \beta_1 m_{t-1} + (1-\beta_1)g_{i,t}\\
\label{adam3}
m_t &= \beta_2 v_{t-1} + (1-\beta_2)g_{i,t}^2
\end{align}
In these equations, $m_t$ and $v_t$ are the first and second moments, respectively. Both are estimated as $\hat{m_t}=\frac{m_t}{1-\beta_1^t}$ and $\hat{v_t}=\frac{v_t}{1-\beta_2^t}$.
This approach can handle the non-stationarity of the objective function as can RMSProp, but Adam can also overcome the sparse gradient issue that is a drawback in RMSProp~\cite{kingma2014adam}.

\section{Results}\label{sec:results}

\subsection{Data}\label{subsec:data}
Our document collection had $134$ labels as shown in Table~\ref{tab:Document}.\footnote{WOS dataset is shared at \url{http://archive.ics.uci.edu/index.php}} The target value has two levels, $k_0~\in~\{1,..,7\}$ which are~$k_0 \in \{$ Computer Science, Electrical Engineering, Psychology, Mechanical Engineering, Civil Engineering,~Medical Science, biochemistry$\}$ and children levels of the labels, ~$k_\varrho$, which contain~$\{17, 16, 19,  9, 11, 53,  9\}$ specific topics belonging to $k_0$, respectively. To train and test the baseline methods described in Section~\ref{sec:me_ba} and the new hierarchical document classification methods described in Section~\ref{sec:deep_learn}, we collected data and meta-data on $46,985$ published papers available from the \textit{Web Of Science} \cite{poungpair2013web,reuters2012web}. To automate collection we used Selenium~\cite{munzert2014automated} with \textit{ChoromeDriver}\cite{gundecha2015selenium} for the Chrome web browser. To extract the data from the site we used \textit{Beautiful Soup}~\cite{beautifulsoup}. We specifically extracted the abstract, domain, and keywords of this set of published papers. The text in the abstract is the input for classification while the domain name provides the label for the top level of the hierarchy. The keywords provide the descriptors for the next level in the classification hierarchy. Table~\ref{tab:Document} shows statistics for this collection. For example, Medical Sciences is one of the top-level domain classifications and there are 53 sub-classifications within this domain. There are also over 14k articles or documents within the domain of health sciences in this data set.

\begin{table}[H]
\centering
\caption{Details of the document set used in this paper.}
\label{tab:Document}
\begin{tabular}{|c|c|c|}
\hline
Domain                 & \begin{tabular}[c]{@{}c@{}}Number of \\ Document\end{tabular} & \begin{tabular}[c]{@{}c@{}}Number of \\ Area\end{tabular} \\ \hline
Biochemistry           & 5,687                                                          & 9                                                         \\ \hline
Civil Engineering      & 4,237                                                          & 11                                                        \\ \hline
Computer Science       & 6,514                                                          & 17                                                        \\ \hline
Electrical Engineering & 5,483                                                          & 16                                                        \\ \hline
Medical Sciences       & 14,625                                                         & 53                                                        \\ \hline
Mechanical Engineering & 3,297                                                          & 9                                                         \\ \hline
Psychology             & 7,142                                                          & 19                                                        \\ \hline
\textbf{Total}                 & \textbf{46,985}                                                         & \textbf{134}                                                       \\ \hline
\end{tabular}
\end{table}

We divided the data set into three parts as shown in Table~\ref{ta_data}. Data set $WOS-46985$  is the full data set with 46,985 documents, and data sets $WOS-11967$ and $WOS-5736$ are subsets of this full data set with the number of training and testing documents shown as well as the number of labels or classes in each of the two levels. For dataset $WOS-11967$, each of the seven level-1 classes has five sub-classes. For data set $WOS-5736$, two of the three higher-level classes have four sub-classes and the last high-level class has three sub-classes. We removed all special characters from all three data sets before training and testing.

\begin{table}[H]
\centering
\caption{Details of three data sets used in this paper.}
\label{ta_data}
\begin{tabular}{|c|c|c|c|c|}
\hline
Data Set & Training & Testing & Level 1 & Level 2 \\ \hline
WOS-11967        & 8018     & 3949    & 7         & 35        \\ \hline
WOS-46985        & 31479    & 15506   & 7        & 134       \\ \hline
WOS-5736        & 4588    & 1148    & 3         & 11        \\ \hline
\end{tabular}
\end{table}

\begin{table*}[]
\centering
\caption{HDLTex and Baseline Accuracy of three WOS datasets}
\label{ta_results}
\begin{tabular}{|c|c|c|c|c|c|c|c|c|c|}
\hline
\multirow{2}{*}{}         & \multicolumn{3}{c|}{WOS-11967} & \multicolumn{3}{c|}{WOS-46985}  & \multicolumn{3}{c|}{WOS-5736}\\ \cline{2-10} 
                          & \multicolumn{2}{c|}{Methods} & Accuracy & \multicolumn{2}{c|}{Methods}  & Accuracy   & \multicolumn{2}{c|}{Methods}                  & Accuracy                        \\ \hline
\multirow{6}{*}{Baseline} & \multicolumn{2}{c|}{DNN}                      & 80.02                           & \multicolumn{2}{c|}{DNN}                      & 66.95                           & \multicolumn{2}{c|}{DNN}                      & 86.15                           \\ \cline{2-10} 
                          & \multicolumn{2}{c|}{CNN (Yang el. et. 2016)} & 83.29                           & \multicolumn{2}{c|}{CNN (Yang el. et. 2016)} &                  70.46               & \multicolumn{2}{c|}{CNN (Yang el. et. 20016)} & 88.68                           \\ \cline{2-10} 
                          & \multicolumn{2}{c|}{RNN (Yang el. et. 2016)}                      &      83.96                            & \multicolumn{2}{c|}{RNN (Yang el. et. 2016)}                      &                   72.12             & \multicolumn{2}{c|}{RNN (Yang el. et. 2016)}                      &             89.46                    \\ \cline{2-10} 
                          & \multicolumn{2}{c|}{NBC}                      & 68.8                            & \multicolumn{2}{c|}{NBC}                      & 46.2                            & \multicolumn{2}{c|}{NBC}                      & 78.14                           \\ \cline{2-10} 
                          & \multicolumn{2}{c|}{SVM (Zhang el. et. 2008)}                      & 80.65                           & \multicolumn{2}{c|}{SVM (Zhang el. et. 2008)}                      & 67.56                           & \multicolumn{2}{c|}{SVM (Zhang el. et. 2008)}                      & 85.54                           \\ \cline{2-10} 
                          & \multicolumn{2}{c|}{SVM (Chen el et. 2016)}                      & 83.16                           & \multicolumn{2}{c|}{SVM (Chen el et. 2016)}                      & 70.22                           & \multicolumn{2}{c|}{SVM (Chen el et. 2016)}                      & 88.24                           
                          \\ \cline{2-10} 
                          & \multicolumn{2}{c|}{Stacking SVM }                      & 79.45                           & \multicolumn{2}{c|}{Stacking SVM }                      & 71.81                           & \multicolumn{2}{c|}{Stacking SVM}                      & 85.68                           \\ \hline

\multirow{18}{*}{HDLTex}  & ~~~~DNN~~~~                   & DNN                  & \multirow{2}{*}{83.73}          & ~~~~DNN~~~~                    & DNN                  & \multirow{2}{*}{70.10}          & ~~~~DNN~~~~                    & DNN                  & \multirow{2}{*}{88.37}          \\ \cline{2-3} \cline{5-6} \cline{8-9}
                          & 91.43                  & 91.58                &                                 & 87.31                  & 80.29                &                                 & 97.97                  & 90.21                &                                 \\ \cline{2-10} 
                          & DNN                    & CNN                  & \multirow{2}{*}{83.32}          & DNN                    & CNN                  & \multirow{2}{*}{71.90}          & DNN                    & CNN                  & \multirow{2}{*}{90.47}          \\ \cline{2-3} \cline{5-6} \cline{8-9}
                          & 91.43                  & 91.12                &                                 & 87.31                  & 82.35                &                                 & 97.97                  & 92.34                &                                 \\ \cline{2-10} 
                          & DNN                    & RNN                  & \multirow{2}{*}{81.58}          & DNN                    & RNN                  & \multirow{2}{*}{73.92}          & DNN                    & RNN                  & \multirow{2}{*}{88.42}          \\ \cline{2-3} \cline{5-6} \cline{8-9}
                          & 91.43                  & 89.23                &                                 & 87.31                  & 84.66                &                                 & 97.97                  & 90.25                &                                 \\ \cline{2-10} 
                          & CNN                    & DNN                  & \multirow{2}{*}{85.65}          & CNN                    & DNN                  & \multirow{2}{*}{71.20}          & CNN                    & DNN                  & \multirow{2}{*}{88.83}          \\ \cline{2-3} \cline{5-6} \cline{8-9}
                          & 93.52                  & 91.58                &                                 & 88.67                  & 80.29                &                                 & 98.47                  & 90.21                &                                 \\ \cline{2-10} 
                          & CNN                    & CNN                  & \multirow{2}{*}{85.23}          & CNN                    & CNN                  & \multirow{2}{*}{73.02} & CNN                    & CNN                  & \multirow{2}{*}{\textbf{90.93}} \\ \cline{2-3} \cline{5-6} \cline{8-9}
                          & 93.52                  & 91.12                &                                 & 88.67                  & 82.35                &                                 & 98.47                  & 92.34                &                                 \\ \cline{2-10} 
                          & CNN                    & RNN                  & \multirow{2}{*}{83.45}          & CNN                    & RNN                  & \multirow{2}{*}{75.07}          & CNN                    & RNN                  & \multirow{2}{*}{88.87}          \\ \cline{2-3} \cline{5-6} \cline{8-9}
                          & 93.52                  & 89.23                &                                 & 88.67                  & 84.66                &                                 & 98.47                  & 90.25                &                                 \\ \cline{2-10} 
                          & RNN                    & DNN                  & \multirow{2}{*}{\textbf{86.07}} & RNN                    & DNN                  & \multirow{2}{*}{72.62}          & RNN                    & DNN                  & \multirow{2}{*}{88.25}          \\ \cline{2-3} \cline{5-6} \cline{8-9}
                          & 93.98                  & 91.58                &                                 & 90.45                  & 80.29                &                                 & 97.82                  & 90.21                &                                 \\ \cline{2-10} 
                          & RNN                    & CNN                  & \multirow{2}{*}{85.63}          & RNN                    & CNN                  & \multirow{2}{*}{74.46}           & RNN                    & CNN                  & \multirow{2}{*}{90.33}          \\ \cline{2-3} \cline{5-6} \cline{8-9}
                          & 93.98                  & 91.12                &                                 & 90.45                  & 82.35                &                                 & 97.82                  & 92.34                &                                 \\ \cline{2-10} 
                          & RNN                    & RNN                  & \multirow{2}{*}{83.85}          & RNN                    & RNN                  & \multirow{2}{*}{\textbf{76.58}}       & RNN                    & RNN                  & \multirow{2}{*}{88.28}          \\ \cline{2-3} \cline{5-6} \cline{8-9}
                          & 93.98                  & 89.23                &                                 & 90.45                  & 84.66                &                                 & 97.82                  & 90.25                &                                 \\ \hline
\end{tabular}
\end{table*}

\subsection{Hardware and Implementation}
The following results were obtained using a combination of central processing units~(CPUs) and graphical processing units~(GPUs). The processing was done on a $Xeon~E5-2640~ (2.6 GHz)$ with $32$ cores and $64 GB$ memory, and the GPU cards were $Nvidia~Quadro~K620$ and $Nvidia~Tesla~K20c$. We implemented our approaches in Python using the Compute Unified Device Architecture~(CUDA), which is a parallel computing platform and Application Programming Interface~(API) model created by $Nvidia$. We also used Keras and TensorFlow libraries for creating the neural networks~\cite{abadi2016tensorflow,chollet2015keras}. 
\subsection{Empirical Results}
Table~\ref{ta_results} shows the results from our experiments. The baseline tests compare three conventional document classification approaches (na\"{i}ve Bayes and two versions of SVM) and stacking SVM with three deep learning approaches (DNN, RNN, and CNN). In this set of tests the RNN outperforms the others for all three $WOS$ data sets. CNN performs second-best for three data sets. SVM with term weighting~\cite{chen2016turning} is third for the first two sets while the multi-word approach of~\cite{zhang2008text} is in third place for the third data set. The third data set is the smallest of the three and has the fewest labels so the differences among the three best performers are not large. These results show that overall performance improvement for general document classification is obtainable with deep learning approaches compared to traditional methods. Overall, na\"{i}ve Bayes does much worse than the other methods throughout these tests. As for the tests of classifying these documents within a hierarchy, the HDLTex approaches with stacked, deep learning architectures clearly provide superior performance. For data set~$WOS-11967$, the best accuracy is obtained by the combination RNN for the first level of classification and DNN for the second level. This gives accuracies of~94\% for the first level, 92\% for the second level and 86\% overall. This is significantly better than all of the others except for the combination of CNN and DNN. For data set~$WOS-46985$ the best scores are again achieved by RNN for level one but this time with RNN for level~2. The closest scores to this are obtained by CNN and RNN in levels~1 and~2, respectively. Finally the simpler data set $WOS-5736$ has a winner in CNN at level~1 and CNN at level~2, but there is little difference between these scores and those obtained by two other HDLTex architectures: DNN with CNN and RNN with CNN.

\section{Conclusions and Future Work}\label{sec:conclusions}
Document classification is an important problem to address, given the growing size of scientific literature and other document sets. When documents are organized hierarchically, multi-class approaches are difficult to apply using traditional supervised learning methods. This paper introduces a new approach to hierarchical document classification, HDLTex, that combines multiple deep learning approaches to produce hierarchical classifications. Testing on a data set of documents obtained from the Web of Science shows that combinations of RNN at the higher level and DNN or CNN at the lower level produced accuracies consistently higher than those obtainable by conventional approaches using na\"{i}ve Bayes or SVM. These results show that deep learning methods can provide improvements for document classification and that they provide flexibility to classify documents within a hierarchy. Hence, they provide extensions over current methods for document classification that only consider the multi-class problem.

The methods described here can improved in multiple ways. Additional training and testing with other hierarchically structured document data sets will continue to identify architectures that work best for these problems. Also, it is possible to extend the hierarchy to more than two levels to capture more of the complexity in the hierarchical classification. For example, if keywords are treated as ordered then the hierarchy continues down multiple levels. HDLTex can also be applied to unlabeled documents, such as those found in news or other media outlets. Scoring here could be performed on small sets using human judges.

\bibliographystyle{IEEEtran}
\bibliography{ref.bib}

\end{document}